\documentclass[10pt, conference, compsocconf]{IEEEtran}

\usepackage[english]{babel}
\usepackage[utf8]{inputenc}
\usepackage{amsmath}
\usepackage{amsthm}
\usepackage{mathtools}

\usepackage{graphicx}
\usepackage{subfig}
\usepackage{floatrow}

\usepackage{float}
\usepackage{algorithm}
\usepackage{algorithmic}

\usepackage{babel}

\newtheorem{definition}{Definition}
\newtheorem{statement}{Statement}

\title{Online Trajectory Segmentation and Summary With Applications to Visualization and Retrieval}

\author{\IEEEauthorblockN{Yehezkel S. Resheff}
\IEEEauthorblockA{Edmond and Lily Safra Center for Brain Sciences \\
The Hebrew University of Jerusalem \\
yehezkel.resheff@mail.huji.ac.il}
}
\begin{document}
\maketitle

\begin{abstract}
Trajectory segmentation is the process of subdividing a trajectory into parts either by grouping points similar with respect to some measure of interest, or by minimizing a global objective function. Here we present a novel online algorithm for segmentation and summary, based on point density along the trajectory, and based on the nature of the naturally occurring structure of intermittent bouts of locomotive and local activity. We show an application to visualization of trajectory datasets, and discuss the use of the summary as an index allowing efficient queries which are otherwise impossible or computationally expensive, over very large datasets. 

\end{abstract}


\section{Introduction}

The steadily dropping cost of data collection and storage is impacting both industry and science. This so-called "data-era" has not skipped movement data, which is now being acquired at unprecedented rates. In the field of Ecology and Animal Tracking, movement data is being collected at a global \cite{rotics2016challenges, harel2016adult} as well as a regional \cite{bijleveld2016understanding, weiser2016characterizing} scale. Car \cite{kullingsjo2012swedish, kim2004imm} and ship \cite{statheros2008autonomous} trajectories are recorded for control, optimization, and safety purposes, and pedestrians are tracked for health, safety, and navigation utilities \cite{foxlin2005pedestrian, alzantot2012uptime}. With this large amount of data rolling in, new and more efficient methods must be developed in order to visualize, analyze, and gain insight and knowledge.

One of the most fundamental computations associated with understanding movement data is segmentation of trajectories. Traditionally, a segmentation proceeds by defining a feature of a single point, and then dividing the entire trajectory into sub-trajectories which are uniform (in some sense) with respect to this feature \cite{buchin2010algorithmic}.  

Features used for this purpose include truly point-wise metrics, such as speed and heading \cite{buchin2010algorithmic}, as well as those designed to capture the behavior in the vicinity of a point, such as first passage time (FPT) \cite{fauchald2003using} and residence time \cite{barraquand2008animal}.

A slightly different approach to trajectory segmentation is based on directly optimizing a cost function related to the segments themselves. Warped K-Means \cite{leiva2013warped} is a trajectory oriented adaptation of the well-known K-Means algorithm, where the regular (mean square distance from centroid) objective is locally optimized, under the additional constraint that a cluster may only contain consecutive points. 

The method we propose here addresses two main limitations of the warped K-Means algorithm. First, for very large datasets it would be necessary to have an online algorithm able to deal with an unknown number of segments. Second, the optimization criterion used should depend on the local trajectory density, allowing long-distance locomotion to be segmented differently from dense local regions. These ideas are formalized in section \ref{section:traj-seg}.

We further present two main applications of the proposed method, addressing major tasks related to large-scale trajectory data. The first is the task of visualization of a large number of possibly dense and overlapping trajectories. The challenge arising in this context is that amounts of data are often prohibitively large and can't be rendered simply on a map. Even if all points are plotted, the trajectories will be obscured by the sheer amount of points. We show that our method is appropriate for descriptive visualization of a large number of trajectories, and allows single paths to be easily resolved and compared.

The second application which we briefly discuss is fast querying and retrieval over a trajectory database. Here, the proposed method serves as an index into the full trajectory data, allowing various otherwise intractable queries. Combined with the visualization component, we suggest a real-time visual search engine for exploration of large scale trajectory data. 

Our main contributions are the online segmentation algorithm, which is a fundamental pre-processing step and may be used for a wide range of downstream processing, and the visualization method based on it, allowing concurrent visualization of many trajectories. The query engine initially described here will be further developed in future research.

The rest of the paper is structured as follows: section \ref{sec:related} briefly reviews related work in trajectory segmentation. In section \ref{section:traj-seg}, the algorithm is described and discussed. Finally, section \ref{sectoin:appliations} contains the proposed applications of the segmentation method for visualization and data retrieval.

\section{Related Work}
\label{sec:related}

In the field of Ecology, Trajectory segmentation is widely used  in order to segment the path of an animal into functionally homogeneous units. The main approach used is to compute a single point-wise feature along the trajectory and then group similar points, with respect to the feature. For example, a method of change-point analysis \cite{lavielle2005using} has been used in conjunction with Residence Time (a metric of the total amount of time spent in the vicinity of a point \cite{barraquand2008animal}). Another approach is to segment a trajectory with respect to the momentary behavior of an animal along a path \cite{resheff2014accelerater}.

A more general framework was also suggested \cite{buchin2010algorithmic}, based on finding the segmentation with the minimum number of segments, such that a given metric will not differ within any segment by more than a pre-defined factor. For a wide range of metrics (such as speed, velocity, heading, curvature, etc.), this can be achieved in $O(nlogn)$.

Warped k-means \cite{leiva2013warped} adopts a completely different view to the segmentation problem. Since this is essentially a problem of clustering similar points, the method attempts (as in the well known K-means algorithm \cite{lloyd1982least}) to find centroids and assignments, in order to minimize the mean square distance of each point to the centroid it is assigned to, under the additional constraint that if two points are in the same cluster, so are all the points between them along the trajectory. 

As the volume of information increases, traditional methods which were very useful for small to medium datasets are no longer applicable, and fast (preferably linear time, online methods) must be developed. In the next section we start by introducing definitions and notation, then describe the proposed algorithm. Then in section \ref{sectoin:appliations} we continues with two applications of the method, which are fundamental to trajectory analysis. The first application we demonstrate is concurrent visualization of many trajectories. Next, we discuss the application of the method to the problem of data retrieval over large datasets.   

\begin{figure}
\centering

\subfloat[]{\includegraphics[width=\textwidth,trim={6.9cm 4.4cm 3cm 2.5cm},clip]{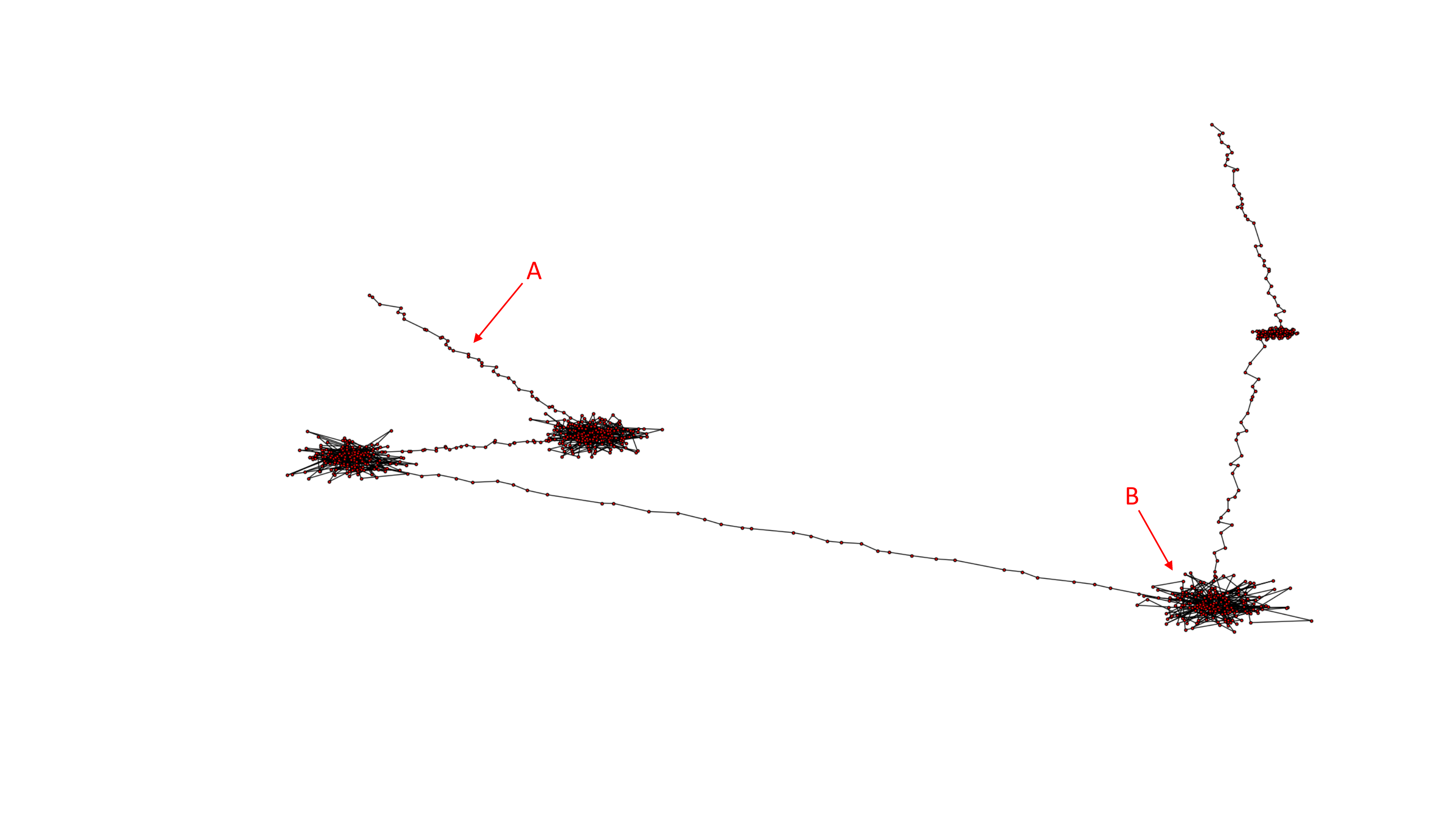}}

\subfloat[]{\includegraphics[width=\textwidth,trim={7cm 3.5cm 3.5cm 2.9cm},clip]{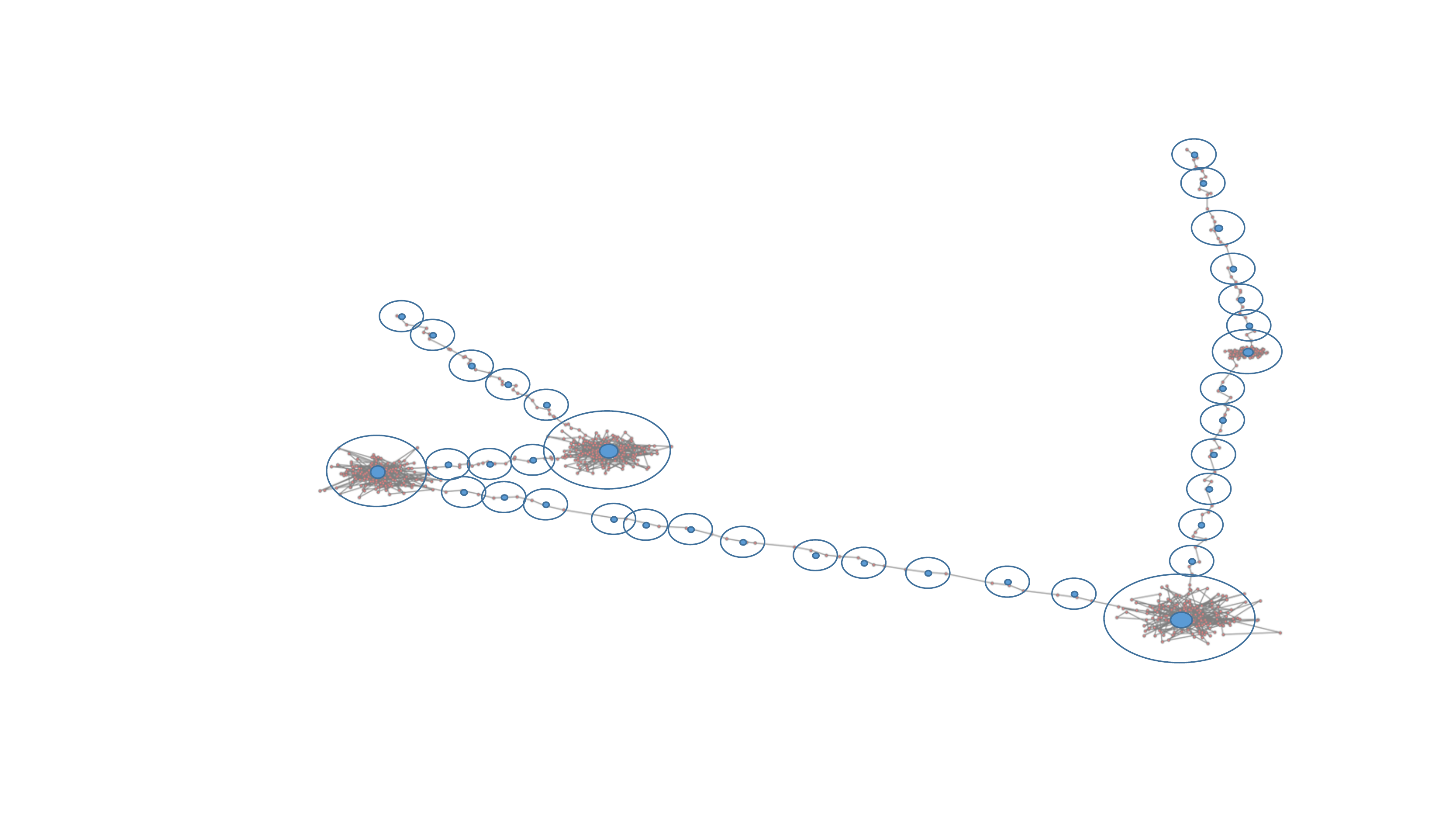}}

\subfloat[]{\includegraphics[width=\textwidth,trim={7cm 3.8cm 4.2cm 2.5cm},clip]{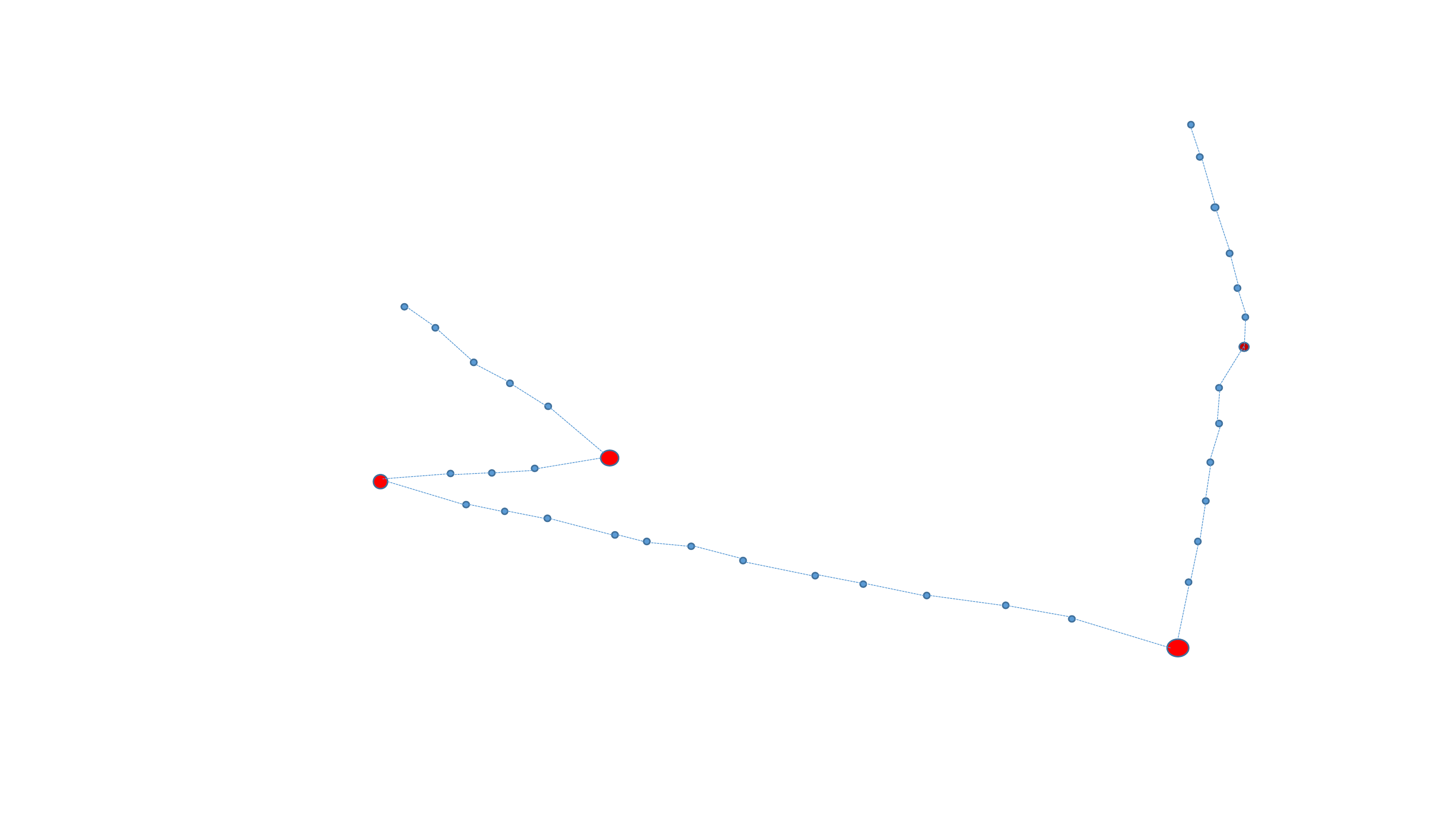}}

\caption{An outline of the segmentation method. (a) A raw trajectory consisting of locomotive parts (a-A) and local parts (a-B). (b) A segmentation overlayed over the raw trajectory. During locomotive periods, a new point is added to the segmentation only when a pre-defined distance is crossed from the previous point. Local periods are represented by a single point (c) the centroids of the segments form a condensed representation of the trajectory, which is in itself a trajectory. Red points correspond to local parts of the original trajectory. }
\label{fig:trajectory}
\end{figure}

\section{Trajectory Segmentation}
\label{section:traj-seg}

\begin{definition}
A trajectory $T$ is a continuous mapping from a time interval to positions:
\begin{equation*}
T: [a,b] \rightarrow R^k
\end{equation*}
\noindent where $T(t)$ is the position at time $t \in [a,b]$
\end{definition}

\noindent in most cases, since we are dealing with trajectories in the physical world, we will have $k \in \{2,3\}$. While all data presented in this paper is $2D$ trajectory data, the concepts and methods discussed are valid for trajectory in arbitrary dimensional space. 

\begin{definition}
A sampled trajectory is a set of time and position tuples:
\begin{equation*}
\{(t_i, p_i)\}_{i=1}^{n}
\end{equation*}
\noindent where for some trajectory T, and some set of timestamps $t \in [a,b]$, $p = T(t)$. 
\end{definition}

In practice, almost all trajectory data is obtained in the form of a sampled trajectory. We note that this notation allows for either equally or unequally spaced samples. We will always assume that the samples are sorted by time, ascending.  

\begin{definition}
\label{def-segmentation}
A segmentation of a sampled trajectory $\{(t_i, p_i)\}_{i=1}^{p}$ is a list of cutoff indexes:
\begin{equation*}
1 = c_0 < c_1 < ... < c_k = p
\end{equation*}
\noindent The $i-th$ segment is then a subset of the sampled trajectory:
\begin{equation*}
 \{(t_i, p_i)\}_{i=c_{i-1}}^{c_i}
\end{equation*}
\noindent  Each segment, is itself a sampled trajectory. 
\end{definition}

\begin{algorithm}
 	input: 
 	\begin{itemize}
 		\item $T$ - sampled trajectory 
 		\item $min\_r$- minimal radius for a segment
 		\item $min\_density$ - minimal density for a segment or radius larger than $min\_r$
 	\end{itemize}
 	output:
 	\begin{itemize}
 		\item $cutoffs$ - the cutoff values for the segmentation of $T$
        \item $centroids$ - the list of centroids of each segment. By adding a time-stamp this becomes a sampled trajectory. Omitted for simplicity. \\	

	\end{itemize}
 	
 	\begin{algorithmic}[1] 
 		
 		\STATE{$cutoffs \leftarrow $ empty list }
        \STATE{append $0$ to $cutoffs$ }
        
        \STATE{$centroids \leftarrow $ empty list } 		
 		\STATE{$current\_centroid \leftarrow T[0] $}
        
        \STATE{$n\_points \leftarrow 1 $}
        \STATE{$radius \leftarrow 0 $}
        
 		\FOR{i = 1 to T.length}
        	\STATE{$n\_points \leftarrow n\_points + 1$}
            \STATE{$radius \leftarrow max(radius, dist(current\_centroid, T[i])) $}
                        
	 		\IF {$radius > min\_r$}
            	
                \STATE{$density \leftarrow n\_points / (pi*radius^2)$}
                
		 		\IF{density $ < min\_density$}     
			 		\STATE{append $i$ to $cutoffs$}
                    \STATE{append $current\_centroid$ to $centroids$}
			 		\STATE{$current\_centroid \leftarrow T[i]$} 
                    \STATE{$n\_points \leftarrow 1$}
                    \STATE{$radius \leftarrow 0 $}
                    \STATE{\textbf{continue}}
		 		\ENDIF
                
	 		\ENDIF
            
            \STATE{$current\_centroid \leftarrow ((n-1)*current\_centroid + T[i]) / n $}
 		\ENDFOR
        
        \STATE{append $T.length$ to $cutoffs$ }
        \STATE{append $current\_centroid$ to $centroids$}
		\STATE{\textbf{return} $cutoffs, centroids$}		 	
 	\end{algorithmic}
 	
 	\protect\caption{Trajectory Segmentation}
 	\label{alg:1}
 \end{algorithm}
 
\subsection{Algorithm outline }

The proposed method is based on the observation that for many types of trajectories, there exist intermittent periods of dense and sparse positions (Figure \ref{fig:trajectory}a). Consider for instance the trajectory of a person throughout a day. Dense local segments correspond to time spent at and around the home, workplace, shopping mall etc. (sometimes called a stay-point \cite{zheng2009mining, li2008mining, zheng2010geolife}), whereas between these segments we might expect long and sparsely distributed locomotive activity. The same pattern is seen for animals; foraging behavior is characterized by dense usage of relatively small areas, whereas long locomotive segments connect between such areas. 

The spread of points during a period of local activity has more than one source. When measuring the position of a completely static entity, measurement noise will cause the locations to be presented as a cloud around the actual location. While this in itself can be overcome by averaging (neglecting adverse boundary effects depending on the length of the filter), for a locally moving entity, the measured spread of points will be an a sum of the actual movement and the noise. Furthermore, when operating on a stream, it is not clear how to average local periods while leaving locomotive periods intact. 

The segmentation approach allows differential consideration of local (or static) versus locomotion parts of a trajectory. The proposed segmentation algorithm will average a full local segment, reducing it to a single centroid, while maintaining much of the structure of the other parts of the trajectory (up to a pre-defined radius).  

\noindent When constructing the algorithm, we require the following:

\begin{enumerate}
	\item The segmentation must work on a trajectory stream (i.e. $O(n)$ and finite memory)
    \item For a given trajectory, the segmentation should be invariant to sampling density. For instance, for a trajectory derived from a drive from home to work, whether it is a fast drive, or a constant traffic jam (corresponding to dense or sparse sampling in space, assuming a constant sampling rate in time), the size of the resulting segmentation should be the same. 
    \item Dense regions corresponding to local activity (shopping area for people, foraging for animals), should be placed in a single segment irrespective of time or area.  
\end{enumerate}

In order to satisfy the first requirement, the algorithm (Algorithm \ref{alg:1}) proceeds in one pass over each trajectory. The number of points, and the radius and density of the current segment are constantly maintained. The second requirement is satisfied by starting a new segment only after a pre-defined radius is passed (i.e. there is a minimum radius for a segment). As a result, denser sampling will have no effect on the segments produced. 

The third requirement is satisfied by letting a segment grow past the pre-defined radius as long as the density in the segment is large enough. See Algorithm \ref{alg:1} for the full account of the method, and Figure \ref{fig:trajectory}b-c for a visualization.   

The output of Algorithm \ref{alg:1} is both the segmentation and the centroids of each of the segments. Recall a segmentation (definition \ref{def-segmentation}) is a list of the cutoff indexes, and thus does not serve in itself as a (standalone) condensed representation of the trajectory. The segment centroids however do serve as such a representation, and the applications we suggest for this method (section \ref{sectoin:appliations} below), are all based on considering them instead of the full data. We note that by adding a time-stamp to the centroids, they become themselves a sampled trajectory. This time-stamp, will in practice be either the mean time of the points in the segment, or the time of the beginning of the segment.   

\subsection{Comments on the calculation of density of points in a segment}

A central component of Algorithm \ref{alg:1} is maintaining of the measure of density of points in the current segment of the trajectory. This is done by maintaining the centroid and radius of the segment -- effectively, the smallest circle (or sphere if the trajectory is in higher dimension), in which the segment is contained. The density is then the number of points divided by the area (Algorithm \ref{alg:1}; line 11). 

The limitation of this approach is that when the segment is not a in the shape of a circle (as is often the case with trajectory data), the circle may highly over-estimate the volume in which the points are actually contained. In this section we develop the idea of replacing the circle with other means of calculation, to better estimate the density, while maintaining the linear runtime and bounded memory characteristics of the original algorithm. 

\begin{algorithm}
	input:
    \begin{itemize}
    	\item stream - sampled trajectory in the form of a stream \\
    \end{itemize} 
    
 	\begin{algorithmic}[1] 
 	
    \STATE {$maxx, maxy \leftarrow -\infty$}
    \STATE {$minx, miny \leftarrow +\infty$}
    \STATE {$n \leftarrow 0$}
    
 	\WHILE {stream.has\_next()}
    	\STATE {$x, y \leftarrow $ stream.next()}
        \STATE {$n \leftarrow n + 1$}
        \STATE {$maxx \leftarrow max(maxx, x)$}
        \STATE {$maxy \leftarrow max(maxy, y)$}
        \STATE {$minx \leftarrow min(minx, x)$}
        \STATE {$miny \leftarrow min(miny, y)$}
        \STATE {$density \leftarrow \frac{n}{(maxx - minx)(maxy - miny)}$}
    \ENDWHILE
       	 	
 	\end{algorithmic}
 	
 	\protect\caption{Online Rectangular Approximation of Point Density}
 	\label{alg:online-rec-approx}
 	
\end{algorithm}
 
 We now formalize the above discussion and first point out that both circles and rectangles are worst-case unbound under-approximators of density. The choice of approximation is dependent therefor on the nature of points expected. We further show that the rectangular approximation can be computed on a stream in linear time and bounded memory, as we require in this setting. Finally, we discuss other more general approaches.    
 
\begin{definition}
The density of a set of points is the number of points divided by the volume of the convex hull of the points
\end{definition}

\begin{statement}
The estimation of point density using the number of points divided by the volume of the smallest (a) sphere or (b) axis-aligned rectangle around the points is an unbound under-estimation.
\end{statement}

\begin{proof}
(part a) Consider the points $(0,0), (a,0), (0, \frac{4}{a}), (a, \frac{4}{a})$ for some $a \in R$. These $4$ points form a rectangle with an area of $4$ and therefor have a density of $1$. The smallest circle around the points has a circumference of at least $a$ (since both the points $(0,0)$ and $(0,a)$ are inside the circle), and therefor an area of at least $\pi\frac{a^2}{4}$. The density estimate using the circle is now at most: 

\[ \frac{4}{\pi\frac{a^2}{4}} = \frac{1}{\pi a^2} \xrightarrow[a\to\infty]{} 0 \]

\noindent meaning the approximation is arbitrarily bad. 

(part b) Consider the points $(\epsilon, 0), (0, \epsilon), (1, 1-\epsilon), (1-\epsilon, 1)$ for some $\epsilon \in R$. These $4$ points are enclosed in an axis-aligned square of area $1$, while forming themselves a rectangle of area:

\[ 1 - (1-\epsilon)^2 -\epsilon^2 \xrightarrow[\epsilon\to0]{} 0 \] 

\end{proof}

\begin{statement}
The rectangular estimation of density can be computed on a stream with $O(n)$ runtime and $O(1)$ memory.
\end{statement}

\begin{proof}
The computation proceeds by considering each point in the stream at a time, and updating the maximal and minimal coordinate encountered on each axis, the total count of points, and the density (See Algorithm \ref{alg:online-rec-approx}).
\end{proof}

A more general approach is to directly compute the convex hull of the points in the segment, and thus derive the density as the ratio of the number of points and the area of the convex hull. Exact convex hull computation is achieved in $O(nlogn)$ (even in the online setting \cite{preparata1985convex}), although linear time approximations exist \cite{hossain2013constructing}.

The downside of this approach is that these methods use $O(n)$ memory. For segments that are expected not to be too long, this may still be feasible. In such cases we are able to compute the density in a more precises manner, by adapting algorithm \ref{alg:1} to use the convex hull computation rather than the circle approximation. 
 
An alternative approach would be to define the density with respect to a cover of the points with a given number of balls, rather than the full convex hull. This area can be approximated using streaming k-means \cite{ailon2009streaming}. 

\subsection{The online algorithm}

The method as described in Algorithm \ref{alg:1} is applied to the entire trajectory (the algorithm is assumed to receive the entire trajectory as input). However, since the algorithm iterates a single time over the points, and keeps only a constant-size summary of the current segment at each point, the adaptation necessary for the application to a stream from many concurrent trajectories is straightforward. 

In the streaming version of the algorithm there needn't be an initialization phase, other than constructing an empty list of known trajectories. The points arriving are considered each at a time, together with a unique ID of the trajectory object they belong to. If the ID is not in the list of known trajectories then it is added, and a new centroid, point counter, and radius are initialized for the new trajectory (as in lines 4-6 of algorithm \ref{alg:1}). Then, the update step (rows 8-17) is conducted using the three stored values belonging to the current trajectory ID. The exceptions are rows 13-14 where the cutoff and centroid are stored to a disk rather then maintained by the process. 

As a result, the amount of memory used by the process is constant (and extremely small) per trajectory, and thus linear in the number of trajectories being processed concurrently. This property makes it feasible to construct a high throughput mechanism when a very large number of trajectories are processed either sequentially or in parallel with shared memory, and then saved both in the raw and segmented formats to separate storage components. This idea is further developed in section \ref{sub-sec:fast-traj-query} below, in the context of data retrieval.     

\subsection{Selecting Hyper-parameters}

The proposed algorithm uses two hyper-parameters (namely the minimal radius of a segment, and minimal density of the large local segments). The choice of the radius parameter effectively controls the granularity of the locomotive parts of trajectory which will be kept in the segmented representation (see Fig. \ref{fig:trajectory}). The selection of this parameter is a trade-off between the size and granularity of the result (the larger the radius, the fewer points we keep, but the less precise the information retained). 

In some cases, it may be beneficial to apply the algorithm in parallel with more than one radius parameter, thus constructing more than one segmented representation at different levels of granularity. Alternatively, and more interestingly, it is possible to iteratively apply the method on the output of itself, thus effectively constructing a hierarchy of segmentations. This is especially applicable when movement has similar structure regardless of scale \cite{rhee2011levy}.

The density parameter is chosen according the the minimal density expected in local segments, and requires some trial and error. When the density parameter is too high, dense local regions will be split into more than one segment, thus increasing the overall size of the output.

In order to compute the total size of the segmentation, it is enough to consider a consecutive locomotive and local region (since they appear intermittently). Consider a length $l$ locomotive part of a trajectory consisting of $n_1$ points, followed by a local region consisting of $n_2$ points. Using a radius $r$ segmentation we will keep at most $\frac{l}{r}$ point out of the locomotive part, and a single point for the local part. In total the ratio between the size of the segmentation and the raw data is $\frac{\frac{d}{r} + 1}{n_1 + n_2}$, which can be made small by choosing a large $r$, as discussed above. 

\section{Applications}
\label{sectoin:appliations}

\subsection{Visualization}

\begin{figure*}
\centering

\begin{floatrow}

\subfloat[Full data; 20 trajectories]{\includegraphics[width=\textwidth,trim={9cm 5cm 9cm 5cm},clip]{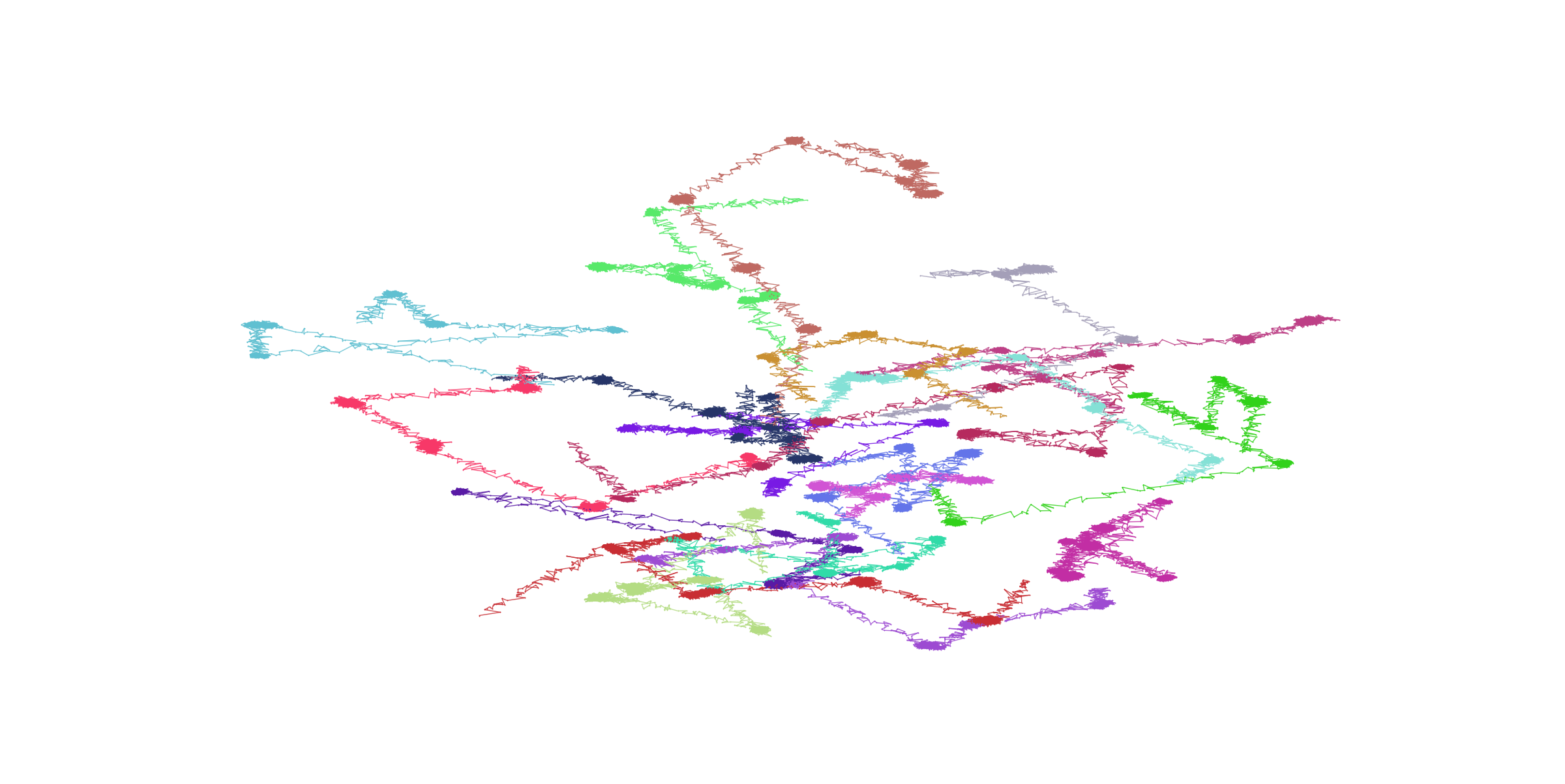}}

\end{floatrow}

\begin{floatrow}

\subfloat[Segmented data; 20 trajectories]{\includegraphics[width=\textwidth,trim={9cm 4cm 7cm 3cm},clip]{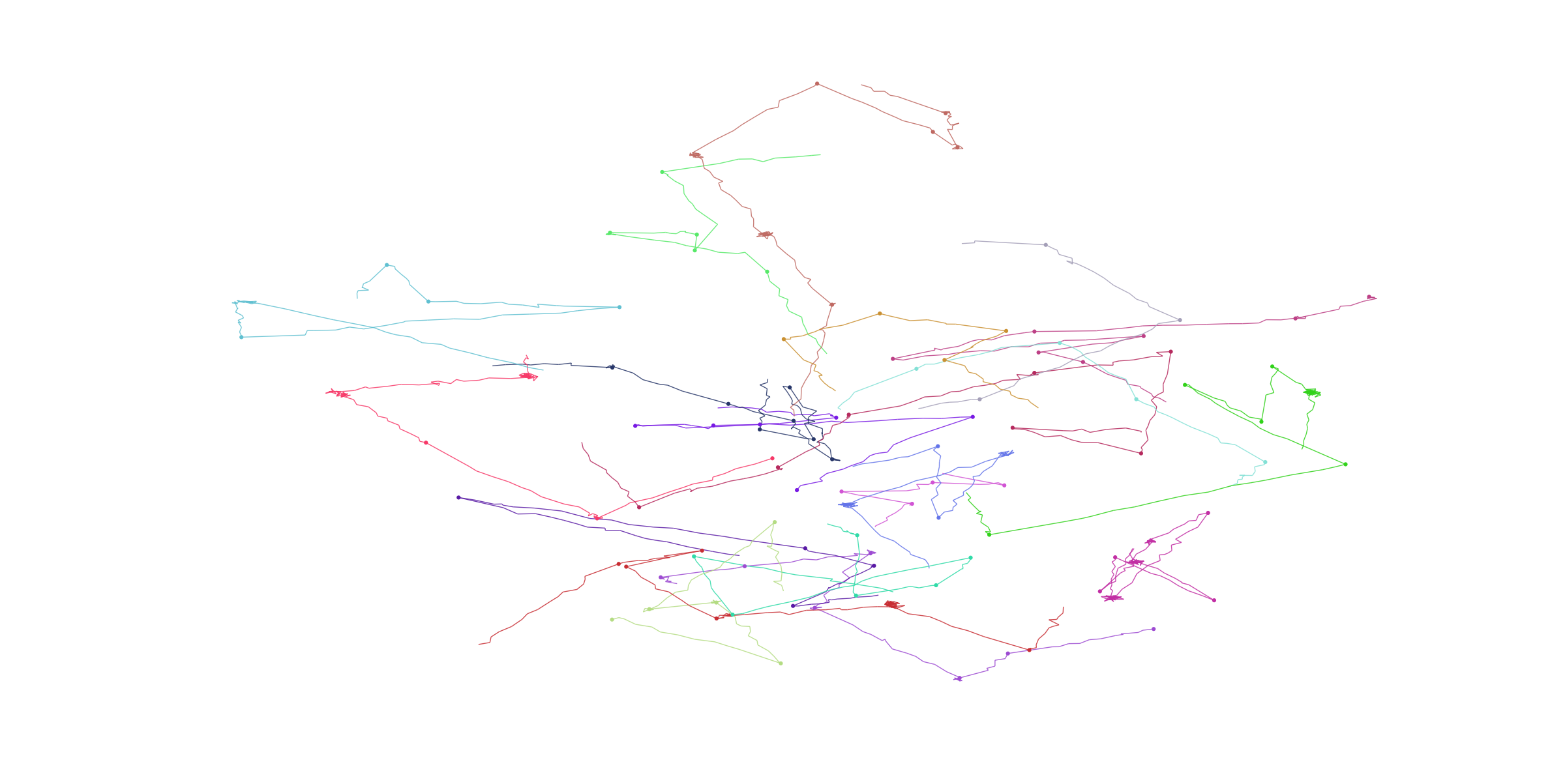}}

\end{floatrow}

\caption{A set of 20 trajectories with a varying number of locomotive and local periods, presented as raw data (a) or the segmented representation (b). }

\label{fig:multiple}
\end{figure*}

Simultaneous visualization of multiple trajectories is especially challenging because of the large amount of data to be displayed all at once. Even when the actual points don't cover the entire display, visual clatter renders individual trajectories indiscernible. 

Several approaches exist for overcoming this problem. In essence, all methods approach the inherent difficulty by presenting a representation of the information \cite{bach2014review}, and differ from each other in the aspect of the data that is to be maintained. 

Trajectory stacking approaches \cite{tominski2012stacking, sun2013survey, andrienko2012visual} are used when all (or most) trajectories in a set are expected to approximately follow a common path. This is a natural state of affairs when dealing with objects moving along a pre-defined path. Such a path could be a road, shipping line, or convenient path for pedestrians. Using the stacking approach, planar trajectories are presented at varying heights on a virtual axis perpendicular to the plane, thus forming a wall-like shape that enables to see the general shape of the common trajectory as well as deviations from it. 

There are two main limitations of the stacking approach. First, directly comparing two trajectories presented at different heights is not visually easy (in order to do this one would have to "imagine" the projection of both onto the plane simultaneously). In fact, to determine the exact position of a trajectory (corresponding to a single projection) is not always easy. 

The second limitation is that while appropriate for visualization of trajectories mostly following a common route, the stacking approach is completely useless for a set of general trajectories in a plane. 

\begin{figure}
\centering

\includegraphics[width=\textwidth,trim={4cm 5.5cm 4cm 4cm},clip]{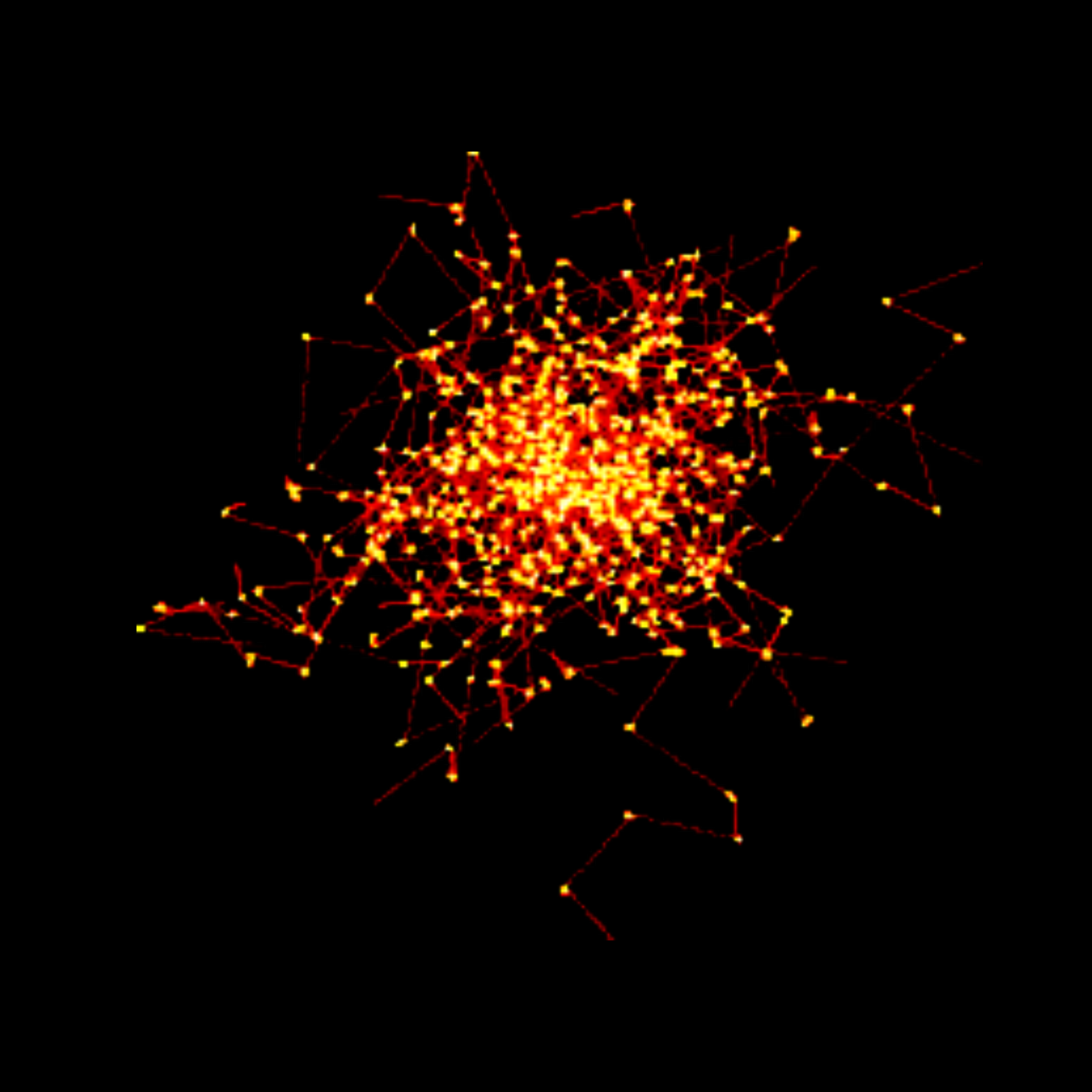}

\caption{A heatmap representation of 100 trajectories. Areas with intense local activity show up while other parts of the trajectory remain invisible. Individual trajectories are not resolvable.}
\label{fig:multiple-trajectory-heatmap}
\end{figure}

Another common visualization approach is the space-use density map (see for example figure \ref{fig:multiple-trajectory-heatmap}). This approach assumes that the pertinent information in a large set of trajectories is the density in space of points pooled over all the set. The result is a visualization as a heatmap presenting the density in a grid. 

The density heatmap approach is mostly useful when single trajectories are of no interest. Clearly, the heatmap does not preserve any information regarding particular trajectories, and thus comparing trajectories is also not possible. 

The assumption made by the visualization method presented here is regarding the nature of the trajectories under consideration, rather than the information needed by the viewer. Namely, we assume that trajectories have the property and consistency of intermittent locomotive and local periods. The locomotive periods are characterized by a directional movement forming a path, while the local periods present a dense use of space, often for extended periods of time. Measurement noise, and small-scale local movement or the order of the noise level turn these periods in to a perceived "cloud" when presenting the full trajectory. 

The proposed method then represents such a trajectory by a decimation in space of the locomotive periods, and a single point for local periods (see algorithm \ref{alg:1}). This representation preserves the entire pertinent information in the trajectory, since locomotive paths are maintained (in the chosen spatial resolution), and local periods are naturally replaced by a single point representing them, without loosing much information.  

The extreme reduction in volume of points to be presented allows simultaneous visualization of many trajectories, while maintaining the ability to resolve single trajectories and compare nearby ones.  

Figure \ref{fig:multiple} presents 20 synthetic trajectories, constructed to have a  varying number of locomotive and local segments, and some overlap between them. When presenting the entire raw data (top panel), single trajectories are not easy to trace and compare to each other. Much of the pertinent information is lost by trajectories occluding each-other, and the general clutter. The segmented representation of the same trajectories (bottom panel) does however expose all the aforementioned information in an intuitive and effortless way. 

In a dynamic setting where an analyst is able to zoom into interesting parts of a trajectory dataset, the segmented representation allows a very large number of trajectories to be examined and compared, in a way not possible with other types of visualization. In the next section we describe a more general storage and analytics mechanism, in which such a visualization capability will serve as one component. 

\subsection{Fast Trajectory Query}
\label{sub-sec:fast-traj-query}

\begin{figure}
\centering
\includegraphics[width=\textwidth,trim={4.5cm 0 4.5cm 0},clip]{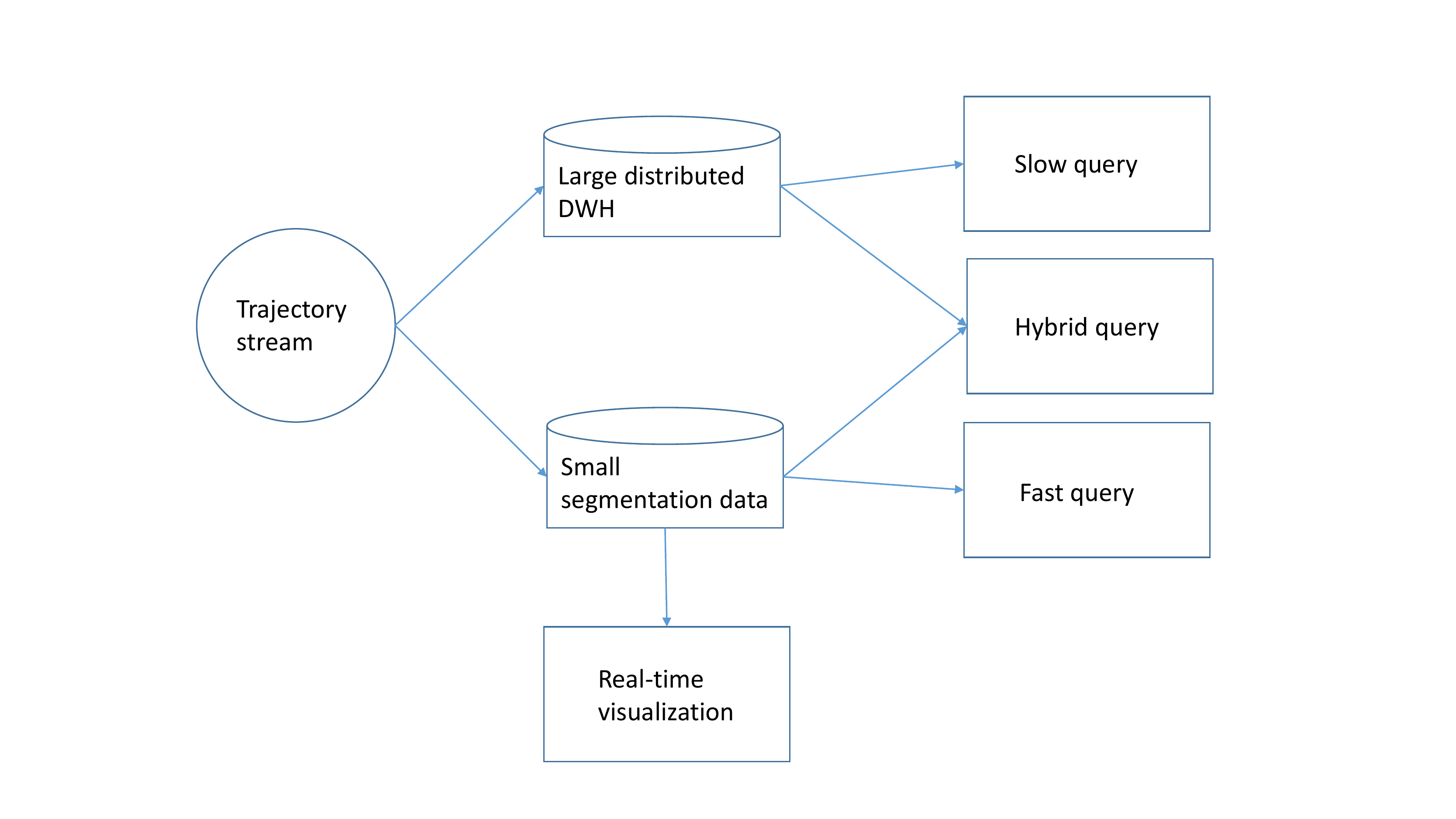}
\caption{The proposed architecture. The trajectory data management system feeds the input stream into a large distributed Data Warehouse (DWH) and also into the small data segmentation representation. Three types of queries are then defined on the bases of the data source they use. Real-time visualization is fed from the small segmentation data. }
\label{fig:diagram}
\end{figure}

With the growing volume of trajectory data collected and stored, the need for new methods of exploring and analyzing such data becomes eminent \cite{bogorny2009st}. Indeed, there has been some recent interest in querying abilities over trajectory data, stemming from multiple domains, as part of the emerging field of trajectory data mining \cite{zheng2015trajectory, zhou2015making, pfoser2000novel}.  

In the field of video analysis, several approaches have been presented for the task of querying using semantics of object trajectories in the video \cite{catarci2003bilvideo, bashir2003segmented, yanagisawa2003shape}. Trajectories embedded in Geographic Information Systems (GIS) have also been addressed, with query methods that take into account a trajectory and its background geographical information \cite{bogorny2009st}.   

So far, none of the proposed methods are able to deal in real-time with truly massive amounts of trajectory data, necessitating large parallel systems. The reason for this is that the size of the index is approximately the size of the raw data. Searching for similarities and patterns over big trajectory data is particularly challenging, since the relevant semantics often stem from large pieces of each single trajectory, which may often be naturally distributed over multiple machines. For this reason, there must be a global index to direct the search.  

The method we propose here is based on the use of compact representation of the segmented trajectories (algorithm \ref{alg:1}) as an index into the full trajectory data, allowing many useful and interesting queries over this small representation. Thus, using segmentation as an index, we are able to convert a hard big-data problem into a tractable small-medium data computation. 

Furthermore, this method allows to preform many trajectory oriented queries without considering actual locations. Using only the segmentation index, without keeping the actual data, may be useful in cases where some computations are necessary, but privacy considerations do not allow storage of full trajectory, or exact location information. (For instance, one could find entities with similar movement patters, or frequent encounters, with only the aggregated segmentations, without any actual exact locations. Another privacy maintaining application would be to determine the context of a user, without storing any locations whatsoever. To this end it would be necessary to assert whether the user is at home, work etc. The details of the method are outside the scope of the current paper.)

The sort of queries for which we can directly use the index comprise most of the typical trajectory based queries:

\begin{enumerate}
	\item Range queries: this sort of query retrieves the identity of entities which were in a certain spatial area, possibly at a certain time. Withing the granularity of the segmentation, this can be preformed without considering the full data.

	\item KNN queries: this common sort of query retrieves the top K trajectories similar in a given metric to an example trajectory. There are many commonly used options for trajectory similarity metrics \cite{zheng2015trajectory}, most of which can be used on the segmented representation. 
    
    \item pairwise queries: many queries over trajectories consider two (or a small set of) trajectories, rather than the entire dataset. Such queries include finding meeting or intersecting points; the point in time where two trajectories come closest to each-other; detecting periods where a small set of trajectories are moving together. All such queries may be computed directly using the segmented representation (where the spatial scale under consideration is no finer than the segmentation radius, otherwise a two-stage method is needed, as is described below). \\    
\end{enumerate}

Another sort of query are the hybrid queries. These queries can be preformed by first narrowing down the search using the segmentation representation, and then diving into the full data only where necessary. For instance, in order to find all trajectories that meet (or intersect) exactly (within a tolerance such smaller than the granularity of the segmentation) with a given trajectory, it is possible to first find all trajectories that pass near the given trajectory, using the course segmentation data. Next, the full data is used only in times and places where such proximity was detected, in order to check whether or nor the trajectories actually meet. When access to a specific trajectory at a specific time in the full data is fast (as is typically the case), this combined method will allow this real-time query in a manner that otherwise may not possible. 

A data management system supporting such queries consists of several building blocks (see Figure \ref{fig:diagram} for a schematic of a possible system). First, the input data is received via a stream of locations along a trajectory, for many (or a huge number of) trajectories simultaneously. These locations are processed in parallel into both a main large Data Warehouse (DWH) and a much smaller database of segmentation data, which is derived from the stream (algorithm \ref{alg:online-rec-approx}). 

A real-time component then allows visualization and fast querying on the small segmentation data. An additional retrieval layer preforms queries either on the full data (slow queries) or hybrid queries where the segmentation data is able to drastically narrow the search in the full data. We defer the full characterization and testing of the system to future research.

\section{Conclusion}

In this paper we describe an online algorithm for segmentation and summary of trajectory data. We then demonstrate the application of the method to visualization of parts of large trajectory datasets, and discuss the potential use as an index into big trajectory data allowing fast queries. 

Visualization of large trajectory datasets is particularly challenging because of the need to represent simultaneously a tremendous number of data-points, in a way the allows single trajectories to be resolved on the one hand, and the big picture of all movement to be discernible of the other hand. Previous methods used either stacked representations which are suitable for the description of many trajectories on the same approximate route (such as a shipping line), or space-use-density based visualizations which are appropriate for describing aggregate behavior over many trajectories. 

The novelty in the proposed approach is in the generation of trajectory summaries, allowing visualization of all the pertinent information across many trajectories at once. The resulting visualization keeps the identity of single trajectories clear and comparable, while at the same time presenting the entire movement in the dataset in a way that is accessible at first glance.

The second application we discuss for the segmentation algorithm is in the field of data retrieval, for indexing of large trajectory data for fast queries. Thus far, the literature contains several trajectory specific indexing and querying frameworks with varying degrees of complexity. However, the size of the index produced is similar to the size of the raw data. The method we propose here utilizes the segmented representation of the trajectory data as a small index allowing fast queries that are otherwise prohibitively slow over large datasets.     

Together with the proposed visualization, we then describe an end-to-end system for ingestion, retention, and real-time visualization and analysis over big trajectory data. Such a system will allow an analyst to utilize both real-time and historical data. Future research will concentrate on building such a system with direct applications to research in Movement Ecology, traffic, and pedestrian movement. 

We note that in addition to the applications of the segmentation method for visualization and data retrieval, the method may be of interest for the sake of the segmentation itself. In many trajectory oriented computations, this is a first step which serves to reduce noise and data volume. The proposed online method is well suited for this purpose, especially when the volume of the data prohibits retention of large parts at the same time, and thus a method must be applied on the stream.  

For example, downstream processing may require differential treatment of locomotive and local segments of trajectories. This is desirably the case for smoothing, where when used on the entire raw data, local segments will tend to get smudged along the locomotive segments leading to and from them. This adverse effect if overcome when preforming the smoothing only on the locomotive segments.  

Another target for future research is the combination of data-structure based trajectory indexing with the current approach. Since the segmentation is in itself a (albeit much smaller) trajectory, methods for efficient representation of complete trajectories which are not suitable for the large-scale data at hand, can be applied to the small segmentation trajectory in order to further improve the approach. In this sense, using the segmentation as an index may serve as a a general framework which can be used in synergy with most existing methods.

\bibliographystyle{plain}
\bibliography{lib}

\end{document}